\documentclass[letterpaper]{article} % DO NOT CHANGE THIS
\usepackage{aaai2026}  % DO NOT CHANGE THIS
\usepackage{times}  % DO NOT CHANGE THIS
\usepackage{helvet}  % DO NOT CHANGE THIS
\usepackage{courier}  % DO NOT CHANGE THIS
\usepackage[hyphens]{url}  % DO NOT CHANGE THIS
\usepackage{graphicx} % DO NOT CHANGE THIS
\usepackage{natbib}  % DO NOT CHANGE THIS AND DO NOT ADD ANY OPTIONS TO IT
\usepackage{caption} % DO NOT CHANGE THIS AND DO NOT ADD ANY OPTIONS TO IT
\usepackage{algorithm}
\usepackage{algorithmic}
\usepackage{amsmath}
\usepackage{amsfonts}
\usepackage{booktabs}
\definecolor{mygray}{gray}{.9}
\usepackage[table,xcdraw]{xcolor}

\usepackage{newfloat}
\usepackage{listings}
\DeclareCaptionStyle{ruled}{labelfont=normalfont,labelsep=colon,strut=off} % DO NOT CHANGE THIS
\floatstyle{ruled}
\newfloat{listing}{tb}{lst}{}
\floatname{listing}{Listing}
\title{Doubly Debiased Test-Time Prompt Tuning for Vision-Language Models}
\author{
    Fei Song\textsuperscript{\rm 1 \rm 2}\equalcontrib,
    Yi Li\textsuperscript{\rm 1 \rm 2}\equalcontrib,
    Rui Wang\textsuperscript{\rm 1 \rm 2},
    Jiahuan Zhou\textsuperscript{\rm 3},
    Changwen Zheng\textsuperscript{\rm 1 \rm 2},
    Jiangmeng Li\textsuperscript{\rm 1 \rm 2}\thanks{Corresponding author.}
}
\affiliations{
    \textsuperscript{\rm 1}National Key Laboratory of Space Integrated Information System, Institute of Software, Chinese Academy of Sciences\\
    \textsuperscript{\rm 2}University of Chinese Academy of Sciences\\
    \textsuperscript{\rm 3}Wangxuan Institute of Computer Technology, Peking University\\
    \{songfei2022, liyi2022, wangrui, changwen, jiangmeng2019\}@iscas.ac.cn, jiahuanzhou@pku.edu.cn
}

\usepackage{bibentry}
\begin{document}

\maketitle

\begin{abstract}
Test-time prompt tuning for vision-language models has demonstrated impressive generalization capabilities under zero-shot settings. However, tuning the learnable prompts solely based on unlabeled test data may induce prompt optimization bias, ultimately leading to suboptimal performance on downstream tasks. In this work, we analyze the underlying causes of prompt optimization bias from both the model and data perspectives. In terms of the model, the entropy minimization objective typically focuses on reducing the entropy of model predictions while overlooking their correctness. This can result in overconfident yet incorrect outputs, thereby compromising the quality of prompt optimization. On the data side, prompts affected by optimization bias can introduce misalignment between visual and textual modalities, which further aggravates the prompt optimization bias. To this end, we propose a Doubly Debiased Test-Time Prompt Tuning method, abbreviated as D\textsuperscript{2}TPT. Specifically, we first introduce a dynamic retrieval-augmented modulation module that retrieves high-confidence knowledge from a dynamic knowledge base using the test image feature as a query, and uses the retrieved knowledge to modulate the predictions. Guided by the refined predictions, we further develop a reliability-aware prompt optimization module that incorporates a confidence-based weighted ensemble and cross-modal consistency distillation to impose regularization constraints during prompt tuning. Extensive experiments across 15 benchmark datasets involving both natural distribution shifts and cross-datasets generalization demonstrate that D\textsuperscript{2}TPT outperforms baselines, validating its effectiveness in mitigating prompt optimization bias.
\end{abstract}

\begin{links}
    \link{Code}{https://github.com/FF2127/D2TPT}
    % \link{Datasets}{https://aaai.org/example/datasets}
    % \link{Extended version}{https://aaai.org/example/extended-version}
\end{links}

\section{Introduction}

\begin{figure}[t]
\centering
\includegraphics[width=0.48\textwidth]{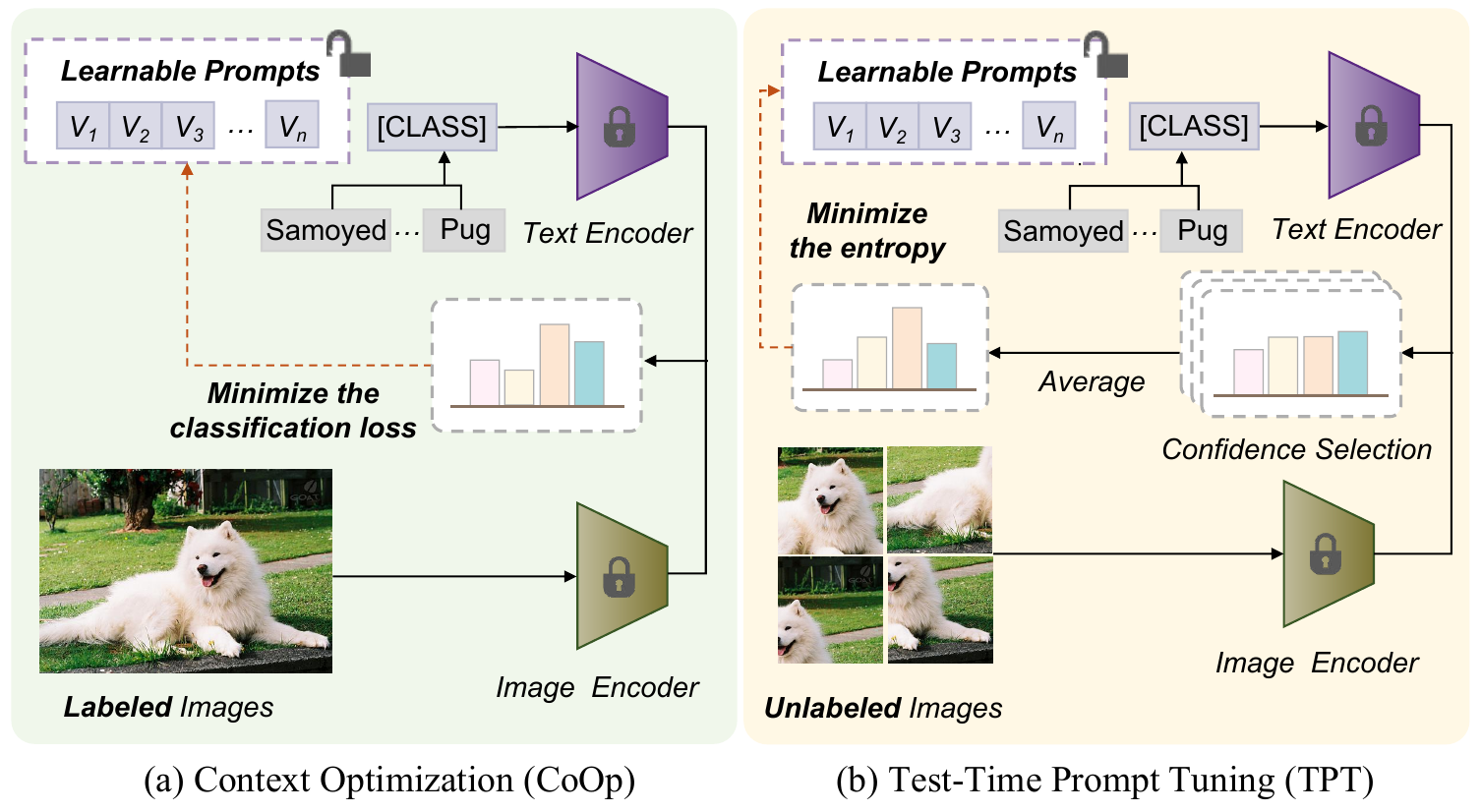} % Reduce the figure size so that it is slightly narrower than the column.
\caption{Comparison of the classical prompt tuning CoOp~\citep{DBLP:journals/ijcv/ZhouYLL22} and test-time prompt tuning TPT~\citep{DBLP:conf/nips/ShuNHYGAX22}. CoOp uses a few labeled samples to optimize the learnable prompt via supervised classification loss, while TPT performs label-free optimization by minimizing the entropy of predictions.}
\label{fig:comparision_coop_tpt}
\end{figure}

Benefiting from large-scale pretraining, vision-language models (VLMs), exemplified by CLIP~\citep{DBLP:conf/icml/RadfordKHRGASAM21}, have demonstrated remarkable zero-shot generalization capabilities across a wide range of downstream tasks~\citep{DBLP:conf/eccv/QianH24,DBLP:conf/eccv/CaoZFCSB24,DBLP:conf/cvpr/ZhangGW024}. However, due to the prevalence of domain shift at test time, vision-language models still suffer from performance degradation when deployed in practical downstream scenarios. To address this issue, prior studies~\citep{DBLP:journals/ijcv/ZhouYLL22,DBLP:conf/cvpr/KhattakR0KK23,DBLP:conf/iclr/RoyE24} have investigated a variety of prompt tuning strategies that adapt the models to downstream tasks using labeled training data. Figure~\ref{fig:comparision_coop_tpt}(a) illustrates a representative method, CoOp~\citep{DBLP:journals/ijcv/ZhouYLL22}, which models the context words of the textual prompt using a set of learnable vectors while keeping the entire pre-trained parameters fixed. Through minimizing the classification loss to optimize these learnable vectors, CoOp achieves superb domain generalization performance. Nevertheless, the reliance on annotated data remains a fundamental bottleneck that hinders the practical deployment of VLMs in real-world applications. 

As a simple yet effective alternative, Test-Time Prompt Tuning (TPT)~\citep{DBLP:conf/nips/ShuNHYGAX22} has attracted widespread attention from researchers in recent years. As shown in Figure~\ref{fig:comparision_coop_tpt}(b), TPT enables prompt tuning without requiring labeled training data by minimizing entropy to adapt the learnable text prompt for each unlabeled test sample. Although test-time prompt tuning methods~\citep{DBLP:conf/iccv/Feng0LKZ23,DBLP:conf/nips/ZhangHZ0L24,DBLP:conf/nips/0009SSX24} have shown promising results in effectively adapting VLMs to target-domain data, relying solely on unlabeled test samples to optimize the learnable prompts is intuitively insufficient and can lead to \textit{prompt optimization bias}. We substantiate this claim by analyzing from both the model and data perspectives. 

\begin{figure}[t]
\centering
\includegraphics[width=0.48\textwidth]{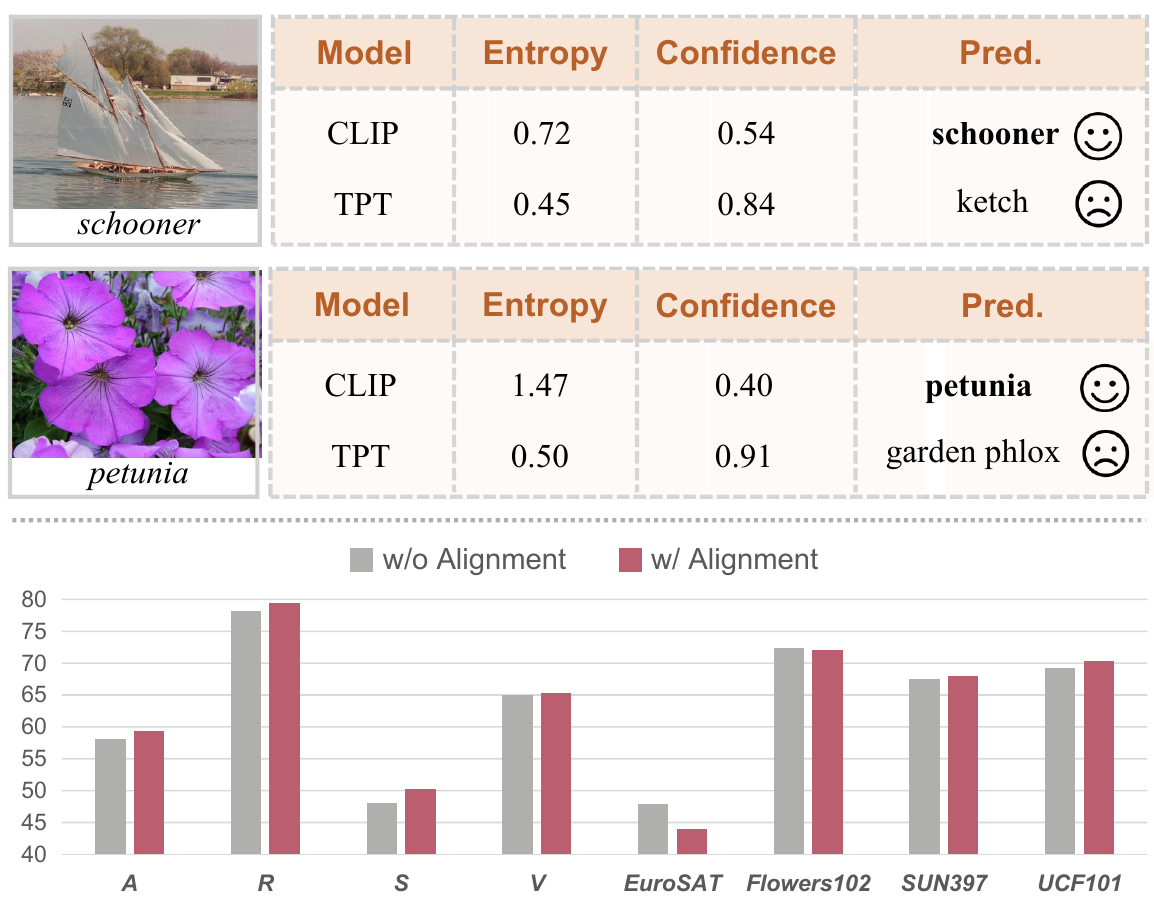} % Reduce the figure size so that it is slightly narrower than the column.
\caption{(Top) Examples illustrating that entropy minimization can lead to overconfident predictions. For instance, although TPT's prediction for petunia has lower entropy and higher confidence, the prediction result is incorrect. (Bottom) Effect of the alignment strategy across different datasets. A, R, S, and V denote the abbreviations of the ImageNet-A, ImageNet-R, ImageNet-Sketch, and ImageNet-V2 datasets, respectively.}
\label{fig:motivation}
\end{figure}

At the model level, existing test-time prompt tuning methods~\citep{DBLP:conf/nips/ShuNHYGAX22} optimize prompts by minimizing the entropy of model predictions on test samples. This objective tends to drive the optimization of learnable prompts based on low-entropy predictions. However, as illustrated in Figure~\ref{fig:motivation}(top), low-entropy predictions are not necessarily correct. When prompt optimization is guided by such incorrect predictions, the model is likely to produce overconfident yet inaccurate outputs. At the data level, incorporating learnable prompts into image-text inputs is, in principle, expected to enhance semantic alignment between the visual and textual modalities within the shared embedding space. Nevertheless, prompts affected by optimization bias may instead deteriorate this alignment. Typical alignment-based test-time prompt tuning ~\citep{DBLP:conf/nips/SamadhGHKNK023} adapts learnable image-text prompts by aligning the statistics of test samples with those of ImageNet. However, this approach is less robust on datasets that differ significantly from ImageNet. As evidenced in Figure~\ref{fig:motivation}(bottom), this alignment strategy performs well only on ImageNet-variant datasets, while yielding limited improvements or even performance degradation on other datasets.

To address the aforementioned issues, we propose a Doubly Debiased Test-Time Prompt Tuning method, abbreviated as D\textsuperscript{2}TPT, which aims to mitigate prompt optimization bias and enable learnable prompts to better support model generalization to downstream tasks. D\textsuperscript{2}TPT consists of two key ingredients: (1) A dynamic retrieval-augmented modulation
module, which introduces a dynamic knowledge base designed to store high-confidence predictions and support continuous updates. When a test image arrives, we use its corresponding feature vector as a query to retrieve the matched class prototype from the knowledge base. The label information associated with the retrieved prototype serves as a high-confidence external supervisory signal to modulate the model's original prediction for the test image. (2) A reliability-aware prompt optimization module, which imposes two regularization constraints on the optimization of learnable prompts, based on the modulated predictions. On one hand, a confidence-based weighted ensemble strategy is designed to integrate discriminative information from augmented views, thereby suppressing the interference caused by low-quality augmentations. On the other hand, we develop a cross-modal consistency distillation strategy, where the image and text modalities alternately act as teacher and student to mutually learn from each other, encouraging semantic consistency between the two modalities within the shared embedding space. The evaluation results on 15 benchmark datasets involving both natural distribution shifts and cross-datasets generalization demonstrate the effectiveness of D\textsuperscript{2}TPT. 

Our key contributions can be summarized as follows:
\begin{itemize}
    \item We reveal that test-time prompt tuning methods solely based on unlabeled test data suffer from prompt optimization bias, and we analyze its underlying causes from both model and data perspectives.
    \item To mitigate the prompt optimization bias, we propose a Doubly Debiased Test-Time Prompt Tuning method, i.e., D\textsuperscript{2}TPT, which consists of a dynamic retrieval-augmented modulation module and a reliability-aware prompt optimization module.
    \item On representative tasks involving natural distribution shifts and cross-datasets generalization, the proposed method consistently improves the model's generalization capability.
\end{itemize}

\section{Related Work}
In this section, we provide an overview of related work on prompting for VLMs and test-time prompt tuning.

\subsection{Prompting for VLMs} 
Leveraging large-scale image-text pairs from the internet, VLMs such as CLIP~\citep{DBLP:conf/icml/RadfordKHRGASAM21} and ALIGN~\citep{DBLP:conf/icml/JiaYXCPPLSLD21} have demonstrated promising performance across a wide range of downstream tasks. Inspired by the success of prompt learning in natural language processing~\citep{DBLP:conf/emnlp/LesterAC21,DBLP:conf/acl/LiL20}, researchers have developed a variety of prompt-based methods to effectively adapt VLMs to downstream tasks using only a few labeled examples. Specifically, CoOp~\citep{DBLP:journals/ijcv/ZhouYLL22} introduces a set of learnable vectors into the input of the text branch and optimizes them by minimizing the classification loss. 
% However, this approach exhibits limited generalization ability in zero-shot settings. To address this issue, CoCoOp~\citep{DBLP:conf/cvpr/ZhouYL022} extends CoOp by learning a lightweight neural network that generates an input-conditional token for each image, which is then combined with the learnable context vectors. 
To further enhance the prompt generalization, Bayesian prompt learning~\citep{DBLP:conf/iccv/DerakhshaniSBCS23} reformulates prompt learning from a Bayesian perspective, casting it as a variational inference problem. Nevertheless, prompting only a single branch of CLIP is sub-optimal, as it limits the flexibility to adapt both representation spaces for downstream tasks. Accordingly, Maple~\citep{DBLP:conf/cvpr/KhattakR0KK23} introduces multi-modal prompt learning across both the vision and language branches to enhance the alignment between their representations. CoPrompt~\citep{DBLP:conf/iclr/RoyE24} imposes a consistency constraint on both the language and vision branches between the trainable and pre-trained models to prevent overfitting on downstream tasks. However, these methods typically rely on labeled training data, thereby limiting their applicability in real-world scenarios.

\subsection{Test-time Prompt Tuning} 
As a simple yet effective approach that adapts vision-language models to downstream tasks at test time without requiring labeled training samples from the target domain, test-time prompt tuning has attracted increasing attention from researchers. TPT~\citep{DBLP:conf/nips/ShuNHYGAX22} first introduces test-time prompt tuning, which optimizes prompts by minimizing entropy to encourage consistent predictions across augmented views. Following this paradigm, a series of subsequent studies have been proposed to further enhance TPT. Specifically, DiffTPT~\citep{DBLP:conf/iccv/Feng0LKZ23} leverages pre-trained diffusion models to generate diverse and informative data, thereby enriching the diversity of augmented views. C-TPT~\citep{DBLP:conf/iclr/YoonYTHLY24} optimizes prompts during test time via enhanced calibration. 
% O-TPT~\citep{DBLP:conf/cvpr/SharifdeenMBKK25} introduces orthogonality constraints on the textual features associated with learnable prompts to further calibrate test-time prompt tuning in VLMs. 
To address the challenge of unsupervised prompt optimization without gradients, B\textsuperscript{2}TPT~\citep{DBLP:conf/aaai/MengCDG25} proposes a black-box test-time prompt tuning that overcomes the gradient-free limitation while reducing complexity. Considering the correlations among test samples, DynaPrompt~\citep{DBLP:conf/iclr/XiaoYHCJHSWS25} builds upon an online prompt buffer to adaptively select and optimize relevant prompts for each test sample during tuning. In this work, we focus on the potential prompt optimization bias that arises when tuning learnable prompts solely based on unlabeled test samples. We analyze its underlying causes from both model and data perspectives and propose targeted strategies to effectively mitigate such bias.
% To improve the robustness of test-time prompt tuning, AdaPrompt~\citep{DBLP:conf/aaai/Zhang0L24} ensembles multiple prompts and employs a confidence-aware buffer to guide prompt adaptation. 

\section{Preliminaries}
Before introducing D\textsuperscript{2}TPT, we first review the preliminaries of test-time prompt tuning for downstream classification tasks using CLIP.

\subsection{Contrastive Language-Image Pretraining (CLIP)}
CLIP is a representative foundation vision-language model (VLM) pre-trained on approximately 400 million image-text pairs. It comprises a visual encoder \( V(\cdot) \) and a textual encoder \( T(\cdot) \), jointly designed to mitigate the semantic gap between visual inputs and textual descriptions. By leveraging a contrastive learning loss, CLIP aligns the visual and textual modalities into a shared embedding space, enabling the learning of generalizable visual representations and enhancing transferability across downstream tasks. For an image classification task with \( C \) categories, CLIP formulates classification via a textual prompt-based approach. Specifically, each class label \( c \in \{1...C\} \) is converted into a text prompt \( \{t_c\}_{c=1}^C \) by prepending a template (e.g., ``a photo of a''). These text prompts are then encoded by the textual encoder to yield the corresponding text feature vectors \( \{\mathbf{z}_{\textit{text}}^{c}\}_{c=1}^C \), where \( \mathbf{z}_{\textit{text}}^{c} = T(t_c) \). Given a test image \( x_{\textit{test}} \), its visual representation is obtained as \( \mathbf{z}_{\textit{img}} = V(x_{\textit{test}}) \). The classification is performed by computing the cosine similarity between the image feature and each class-specific text feature. The resulting class probability distribution is given by:
\begin{equation} \label{eq:clip}
  P(c|x_{\textit{test}}) = \frac{\exp(\textit{cos}(\mathbf{z}_{\textit{img}}, \mathbf{z}_{\textit{text}}^{c}) / \tau_{temp})}{\sum_{c'=1}^{C} \exp(\textit{cos}(\mathbf{z}_{\textit{img}}, \mathbf{z}_{\textit{text}}^{c'}) / \tau_{temp})},
\end{equation}
where \( \textit{cos}(\cdot) \) denotes the cosine similarity function, and \( \tau_{temp} \) is the temperature hyperparameter.

\subsection{Test-time Prompt Tuning for CLIP}
Despite the impressive zero-shot generalization capability of CLIP, effectively adapting it to unseen distributions at test time remains a critical challenge. Furthermore, fine-tuning the entire model is computationally prohibitive, particularly for large-scale transformer architectures. Accordingly, test-time prompt tuning has emerged as a lightweight strategy that adapts the input prompts based solely on the test sample itself, without requiring access to labeled training data. Specifically, given a test image \( x_{\textit{test}} \), TPT generates \(N\) augmented views using a transformation family \(\mathcal{A}\), resulting in \(\{\mathcal{A}_{n}(x_{\textit{test}})\}_{n=1}^{N}\). For each view, CLIP produces a prediction \(P_{\boldsymbol{p}}(c|\mathcal{A}_{n}(x_{\textit{test}}))\) based on a learnable prompt \(\boldsymbol{p}\). The optimization objective aims to minimize the entropy of the average prediction distribution over all augmented views:
\begin{equation} \label{eq:tpt_1}
\boldsymbol{p}^*=\arg\min_{\boldsymbol{p}}\mathcal{H}(\tilde{P}_{\boldsymbol{p}} (c|x_{\textit{test}}))
\end{equation}
% \begin{equation} \label{eq:tpt_1}
%   \boldsymbol{p}^*=\arg\min_{\boldsymbol{p}}-\sum_{c=1}^{C}\tilde{P}_{\boldsymbol{p}} (c|x_{\textit{test}})log\tilde{P}_{\boldsymbol{p}} (c|x_{\textit{test}}),
% \end{equation}
where \(\mathcal{H}(\tilde{P}_{\boldsymbol{p}} (c|x_{\textit{test}}))=-\sum_{c=1}^{C}\tilde{P}_{\boldsymbol{p}} (c|x_{\textit{test}})log\tilde{P}_{\boldsymbol{p}} (c|x_{\textit{test}}) \), and \( \tilde{P}_{\boldsymbol{p}} (c|x_{\textit{test}})=\frac{1}{N} \sum_{n=1}^{N} {P}_{\boldsymbol{p}} (c|\mathcal{A}_{n}(x_{\textit{test}})) \).

To mitigate the impact of unreliable augmentations (e.g., random crops that remove discriminative content), TPT employs a confidence selection mechanism that computes the self-entropy \(\mathcal{H}({P^{n}})\) of each prediction, where \(P^{n} = {P}_{\boldsymbol{p}} (c|\mathcal{A}_{n}(x_{\textit{test}})) \). Then, select only those views with entropy below a threshold \(\tau \), which is determined as the \(\rho \)-percentile among the entropy values of the \(N\) views, ranked from low to high. Thus, the averaged prediction under confidence selection is computed as follows:
\begin{equation} \label{eq:tpt_2}
{\tilde{P}_{\boldsymbol{p}} (c|x_{\textit{test}})}'=\frac{1}{\rho N}\sum_{n=1}^N \mathbb{I}[\mathcal{H}(P^n)\leq\tau]P^{n},
\end{equation}
where \(\mathbb{I}[\mathcal{H}(P^n)\leq\tau]\) denotes an indicator function that returns 1 if the prediction entropy is below the threshold \(\tau \), and 0 otherwise. 

By minimizing the entropy of predictions averaged over selected high-confidence augmentations, TPT adapts prompts in an unsupervised manner, enabling CLIP to better generalize to out-of-distribution data at test time.

\begin{figure*}[t]
\centering
\includegraphics[width=0.95\textwidth]{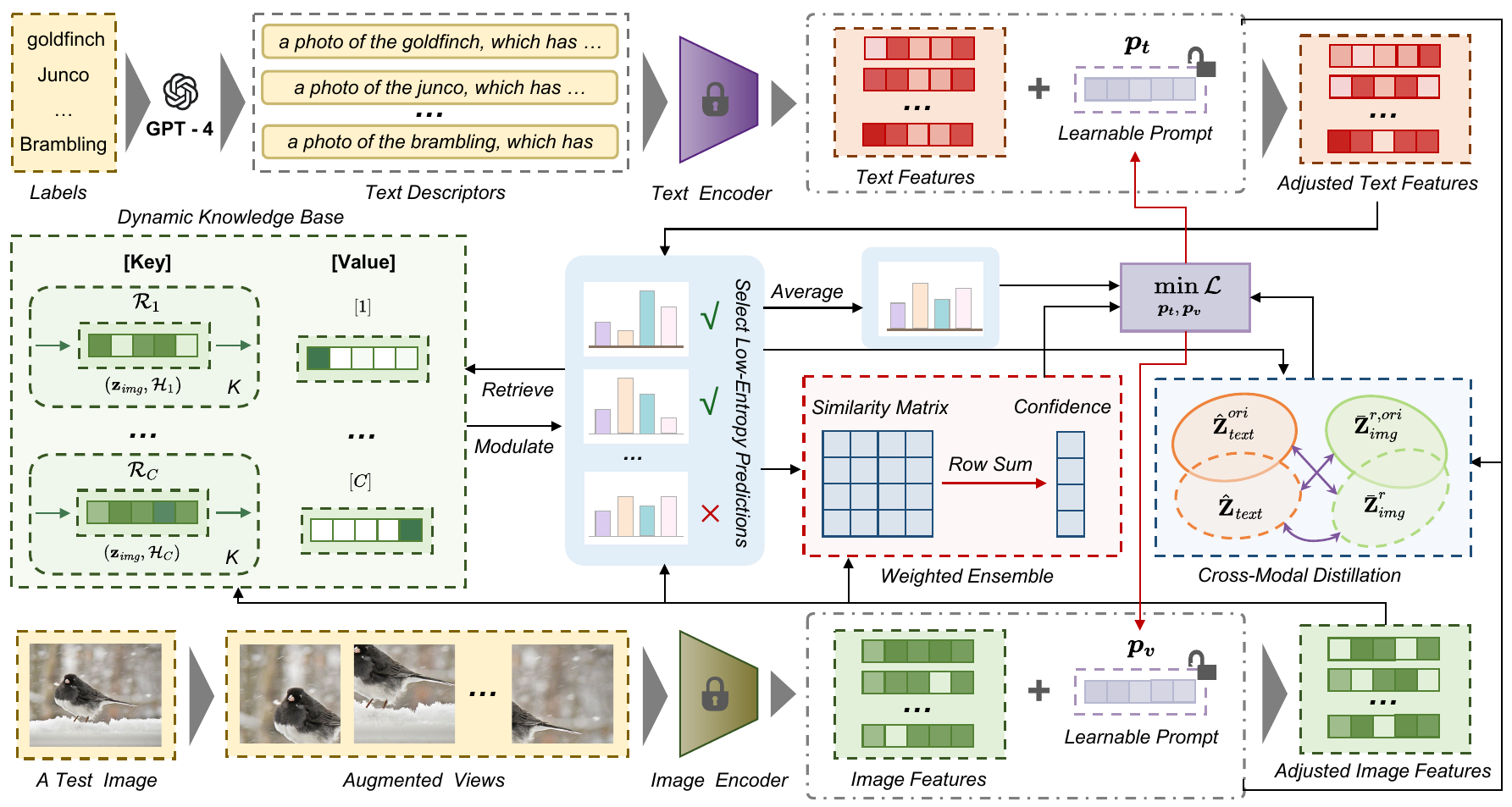} % Reduce the figure size so that it is slightly narrower than the column.
\caption{The overall framework of our D\textsuperscript{2}TPT method.}
\label{fig:framework}
\end{figure*}

\section{Methodology}
In this section, we present our proposed D\textsuperscript{2}TPT method, which introduces modality-specific learnable prompts for image and text inputs, and integrates dedicated modules to jointly mitigate prompt optimization bias. Figure~\ref{fig:framework} illustrates an overview of D\textsuperscript{2}TPT.

\subsection{Modality-Specific Prompt Design}
In this work, we adopt a prompt ensembling strategy that utilizes multiple contextual prompt templates. Specifically, for each class \(c\), general prompt templates \(t_{c}^{\textit{gen}}\) are constructed following CLIP, while class-specific prompt templates \(t_{c}^{\textit{spe}}\) are generated with the assistance of GPT-4~\citep{DBLP:journals/corr/abs-2303-08774}. The textual encoder \( T(\cdot) \) is then applied to obtain the corresponding textual embedding: \( \mathbf{z}_{\textit{gen}}^{c} = T(t_{c}^{\textit{gen}})\) and \( \mathbf{z}_{\textit{spe}}^{c} = T(t_{c}^{\textit{spe}})\). The prototype of these textual prompt templates in the embedding space is calculated as \(\mathbf{z}_{\textit{text}}^{pro, c} = avg(\mathbf{z}_{\textit{gen}}^{c}, \mathbf{z}_{\textit{spe}}^{c})\), where \(avg(\cdot)\) denotes the mean operation. 

Since generating class-specific prompts solely based on class labels can introduce information irrelevant to the input image, we design a learnable textual prompt \(\boldsymbol{p_t}\) that dynamically adjusts each textual prototype to better align with the visual feature of the test image. For each class \(c\), the adjusted textual representation is computed as \({\mathbf{z}_{\textit{text}}^{c}}' = \mathbf{z}_{\textit{text}}^{pro, c} + \boldsymbol{p_t} \). For convenience, we use \(\mathbf{Z}_{\textit{text}}\in \mathbb{R}^{C\times D}\) to represent the learnable text features corresponding to \( C \) class labels, where \(D\) is the feature dimension. 

Similarly, while raw images contain rich semantics, they may also include irrelevant or misleading information that impedes alignment with textual features. Therefore, we also design a learnable image prompt \(\boldsymbol{p_v}\) to adjust the image representation. For each test image \( x_{\textit{test}} \), we generate \(N\) augmented views \(\{\mathcal{A}_{n}(x_{\textit{test}})\}_{n=1}^{N}\) using the transformation family \(\mathcal{A}\), and extract the corresponding features \(\{\mathbf{z}_{\textit{img}}^{n}\}_{n=1}^{N} \) using the visual encoder \( V(\cdot) \), where \(\mathbf{z}_{\textit{img}}^{n}=V(\mathcal{A}_{n}(x_\textit{test} ))\). The adjusted image representation for each view is then computed as \({\mathbf{z}_{\textit{img}}^{n}}' = \mathbf{z}_{\textit{img}}^{n} + \boldsymbol{p_v}\), and we denote the collection of adjusted image features as \(\mathbf{Z}_{\textit{img}}\in \mathbb{R}^{N\times D}\).

\begin{algorithm}[h]
\caption{D\textsuperscript{2}TPT}
\label{alg:d2tpt}
\textbf{Input}: Test image \(x_{\textit{test}}\), a set of categories set \(\{1,\cdots, C\}\), visual encoder \(V(\cdot)\), textual encoder \(T(\cdot)\), learnable prompts \(\boldsymbol{p_v}, \boldsymbol{p_t}\), entropy threshold \(\tau\), register capacity \(K\), hyperparameters \(\alpha\), \(\beta\)\\
\textbf{Output}: Predicted category \(c\)

\begin{algorithmic}[1]
\STATE Adapt visual and textual features with learnable prompts:  \({\mathbf{z}_{\textit{img}}^{n}}' = \mathbf{z}_{\textit{img}}^{n} + \boldsymbol{p_v}\), \({\mathbf{z}_{\textit{text}}^{c}}' = \mathbf{z}_{\textit{text}}^{pro, c} + \boldsymbol{p_t} \);

\STATE Compute similarity logits using normalized vectors: \(\mathbf{L} = \exp{(s)} \cdot\left ( \hat{\mathbf{Z}}_{\textit{img}}(\hat{\mathbf{Z}}_{\textit{text}})^{T}  \right )\);

\STATE Identify high-confidence predictions and pseudo labels using Equations~(\ref{eq:d2tpt_5})–(\ref{eq:d2tpt_6});

\STATE Update class-specific register using visual feature \(\mathbf{z}_{\textit{img}}\), pseudo label \(\check{y}\), and corresponding entropy \(\mathcal{H}_{\check{y}}\) based on Equation~(\ref{eq:d2tpt_9});

\STATE Construct key-value pairs \(\mathbf{K}, \mathbf{V}\) from register features and one-hot labels, and compute retrieval-based logits: \(\mathbf{L}_{R}=\lambda\exp(-\gamma(1-\mathbf{z}_\textit{img}\mathbf{K}^{T}))\mathbf{V}\);

\STATE Fuse \(\mathbf{L}\) and \(\mathbf{L}_{R}\), and compute the corresponding entropy loss as per Equation~(\ref{eq:d2tpt_13});

\STATE Estimate confidence scores for selected images to perform weighted prediction fusion, and compute entropy loss of ensemble logits using Equation~(\ref{eq:d2tpt_15});

\STATE Derive distillation logits from both original and adapted features, and compute the corresponding entropy loss via Equation~(\ref{eq:d2tpt_17});

\STATE Optimize prompts by minimizing total loss defined in Equation(~\ref{eq:L_final}) to produce final prediction \(c\).
\end{algorithmic}
\end{algorithm}

\subsection{Dynamic Retrieval-Augmented Modulation}
Retrieval-Augmented Generation (RAG)~\citep{DBLP:journals/corr/abs-2312-10997} mitigates the limitations of outdated or incomplete model knowledge by retrieving relevant information from external sources and incorporating it into the query. Inspired by this, we propose a dynamic retrieval-augmented modulation module to suppress low-entropy yet incorrect predictions.

Let \(\hat{\mathbf{Z}}_{\textit{text}}\) and \(\hat{\mathbf{Z}}_{\textit{img}}\) denote the normalized textual and visual feature matrices, respectively. The similarity logits \(\mathbf{L} = \exp{(s)} \cdot\left ( \hat{\mathbf{Z}}_{\textit{img}}(\hat{\mathbf{Z}}_{\textit{text}})^{T}  \right )\), where \(s\) is a logit scaling factor and \(\mathbf{L} \in \mathbb{R}^{N \times C}\). To filter out unreliable predictions, we apply entropy-based confidence selection over the logits and obtain a subset of high-confidence predictions:
\begin{equation} \label{eq:d2tpt_5}
\text{L}_{r}=\{\mathbf{L}_{i} \mid i \in \{1,\cdots,N\} \wedge \mathcal{H}(\text{softmax}(\mathbf{L}_{i})) \le \tau \}, 
\end{equation}
where \(\mathbf{L}_{i}=[\text{L}_{i}^{1},\cdots,\text{L}_{i}^{C}] \in \mathbb{R}^{1 \times C}\) denotes the similarity logit for the \(i\)-th augmented view, and \(\text{softmax}(\mathbf{L}_{i}) = \frac{\exp(\mathbf{L}_{i})}{\sum_{c=1}^{C} \exp(\text{L}_{i}^{c})}\). 

Based on the filtered confident predictions, we construct a dynamic knowledge base that stores test image features, pseudo labels, and corresponding entropy values. The predicted labels are determined as:
\begin{equation} \label{eq:d2tpt_6}
Y_{r}=\{\hat{y}_{i} \mid \mathbf{L}_{i}\in\text{L}_{r}  \wedge \hat{y}_{i}=\arg\max_{c}(\text{softmax}(\mathbf{L}_{i}) ) \}, 
\end{equation}
where \(\hat{y}_{i}\) denotes the predicted label corresponding to \(\mathbf{L}_{i}\). 

We identify the most frequently predicted label \(\check{y}=\arg\max_{y \in \{1...C\}} \sum_{i=1}^{\left | Y_r \right | } \mathbb{I}(\hat{y}_{i}=y)\) as the pseudo label, where \(\left | Y_r \right | \) denotes the number of predictions in \(Y_r\). To estimate its confidence, we select the minimum entropy among predictions assigned to \(\check{y}\), computed as \(\mathcal{H}_{\check{y}}=\min_{i\in I_{\check{y}}}\{\mathcal{H}(\text{softmax}(\mathbf{L}_{i}))\}\), where \(I_{\check{y}}=\{i \mid \hat{y}_{i}=\check{y}\}\). 

To maintain a high-quality class-wise knowledge base, we adopt a dynamic updating strategy. Given the visual feature \(\mathbf{z}_\textit{img}\) of the test image \(x_{\textit{test}}\), the pseudo label \(\check{y}\), and the corresponding minimum entropy \(\mathcal{H}_{\check{y}}\), we construct a tuple \(\mathbf{x} = (\mathbf{z}_\textit{img}, \mathcal{H}_{\check{y}})\) for possible inclusion in the register \(\mathcal{R}_{\check{y}}\) associated with the pseudo label \(\check{y}\) in the knowledge base. The register is updated as follows:
\begin{equation} \label{eq:d2tpt_9}
\mathcal{R}_{\check{y}}= 
\begin{cases}
\mathcal{R}_{\check{y}}\cup\{\mathbf{x}\}, & \mathrm{if}0<|\mathcal{R}_{\check{y}}|<K \\
(\mathcal{R}_{\check{y}}\setminus\{\mathbf{x}^{m}\})\cup\{\mathbf{x}\}, & \mathrm{if}|\mathcal{R}_{\check{y}}|= K \wedge \mathcal{H}_{\check{y}}<\mathcal{H}^{m} \\
\{\mathbf{x}\}, & \mathrm{if}|\mathcal{R}_{\check{y}}|=0
\end{cases},
\end{equation}
where \(\mathcal{H}^{m}=\max\{\mathcal{H}_{j}\mid\mathcal{H}_{j}\in\mathcal{R}_{\check{y}}\}\), \(K\) denotes the maximum capacity of \(\mathcal{R}_{\check{y}}\), and \(\mathbf{x}^{m}\) is the tuple corresponding to \(\mathcal{H}^{m}\). The register is sorted by ascending entropy to retain high-confidence samples for reliable test-time retrieval.

During retrieval, key-value pairs are constructed by aggregating the stored visual features for each class. Specifically, the key for class \(\check{y}\in\{1,\cdots,C\}\) is computed as the prototype of visual features in its register, i.e., \(\mathbf{k}_{\check{y}}=\frac{1}{\left |\mathcal{R}_{\check{y}}\right | }\sum_{\mathbf{z}_{i}\in\mathcal{R}_{\check{y}}}{\mathbf{z}_{i}}\in \mathbb{R}^{1\times D}\), and the corresponding value is a one-hot vector \(\mathbf{v}_{\check{y}}=\text{one\_hot}(\check{y})\in \mathbb{R}^{1\times C}\), with 1 at the \(\check{y}\)-th position and 0 elsewhere. Consequently, the key matrix \(\mathbf{K}=\left [\mathbf{k}_{1};\cdots;\mathbf{k}_{\breve{C}}\right ]\in\mathbb{R}^{\breve{C} \times D}\) and value matrix \(\mathbf{V}=\left [\mathbf{v}_{1};\cdots;\mathbf{v}_{\breve{C}}\right ]\in\mathbb{R}^{\breve{C} \times C}\) are formed, where \(\breve{C}\) denotes the number of classes stored in the knowledge base.

Inspired by \citep{DBLP:conf/eccv/ZhangZFGLDQL22}, we compute the retrieval-augmented modulation logits for the test image \(x_{\textit{test}}\) as \(\mathbf{L}_{RAM}=\mathbf{L}+\mathbf{L}_{R}\), where \(\mathbf{L}_{R}=\lambda\exp(-\gamma(1-\mathbf{z}_\textit{img}\mathbf{K}^{T}))\mathbf{V}\) represents the retrieval-based logits, \(\lambda\) is the residual ratio, and \(\gamma\) is the sharpness ratio. We then use Equation (\ref{eq:d2tpt_5}) to select high-confidence logits from \(\mathbf{L}_{RAM}\), forming \(\mathbf{L}_{RAM}^{r}\in\mathbb{R}^{M \times C}\), where \(M\) denotes the number of selected logits. Therefore, the self-entropy loss is computed as:
\begin{equation} \label{eq:d2tpt_13}
\mathcal{L}_{RAM}= \mathcal{H}(\frac{1}{M}\sum_{m=1}^{M} \text{softmax}(\mathbf{L}_{RAM}^{r,m})),
\end{equation}
where \(\mathbf{L}_{RAM}^{r,m}\in\mathbb{R}^{1\times C}\) is the \(m\)-th row of \(\mathbf{L}_{RAM}^{r}\).

% reference R-TPT DPE
\begin{table}
\centering
\small
\setlength{\tabcolsep}{0.7mm}
\begin{tabular}{lcc}
\toprule
Dataset & \#Classes & \#Test \\
\midrule
ImageNet~\citep{DBLP:conf/cvpr/DengDSLL009} & 1,000 & 50,000\\
ImageNet-A~\citep{DBLP:conf/cvpr/HendrycksZBSS21} & 200 & 7,500\\
ImageNet-R~\citep{DBLP:conf/iccv/HendrycksBMKWDD21} & 200 & 30,000\\
ImageNet-Sketch~\citep{DBLP:conf/nips/WangGLX19} & 1,000 & 50,889\\
ImageNet-V2~\citep{DBLP:conf/icml/RechtRSS19} & 1,000 & 10,000\\
\midrule
Caltech101~\citep{DBLP:conf/cvpr/LiFP04} & 100 & 2,465\\
OxfordPets~\citep{DBLP:conf/cvpr/ParkhiVZJ12} & 37 & 3,669\\
StanfordCars~\citep{DBLP:conf/iccvw/Krause0DF13} & 196 & 8,041\\
Flowers102~\citep{DBLP:conf/icvgip/NilsbackZ08} & 102 & 2,463\\
Food101~\citep{DBLP:conf/eccv/BossardGG14} & 101 & 30,300\\
Aircraft~\citep{DBLP:journals/corr/MajiRKBV13} & 100 & 3,333\\
SUN397~\citep{DBLP:conf/cvpr/XiaoHEOT10} & 397 & 19,850 \\
DTD~\citep{DBLP:conf/cvpr/CimpoiMKMV14} & 47 & 1,692\\
EuroSAT~\citep{DBLP:journals/staeors/HelberBDB19} & 10 & 8,100\\
UCF101~\citep{DBLP:journals/corr/abs-1212-0402} & 101 & 3,783 \\
\bottomrule
\end{tabular}
\caption{Overview of datasets used in experiments, including the number of classes and test data.}
% \caption{Overview of datasets used in experiments.}
\label{tab:dataset-overview}
\end{table}

\begin{table*}[t]
\centering
% \adjustbox{max width=\textwidth}{
    \setlength{\tabcolsep}{1.0mm}
    \begin{tabular}{ll ccccccc}
    \toprule
     Method & Publication & ImageNet & ImageNet-A & ImageNet-R & ImageNet-S & ImageNet-V & Average & OOD Average \\
     \midrule
    CLIP-ViT-B/16 & \textit{ICML2021} & 66.73 & 47.87 & 73.98 & 46.09 & 60.86 & 59.11 & 57.20 \\
    \midrule
    TPT & \textit{NeurIPS2022} & 68.98 & 54.77 & 77.06 & 47.94 & 63.45 & 62.44 & 60.81 \\
    DiffTPT & \textit{ICCV2023} & 70.30 & 55.68 & 75.00 & 46.80 & 65.10 & 62.58& 60.65 \\
    PromptAlign & \textit{NeurIPS2023} & - & 59.37 & 79.33 & 50.23 & 65.29 & - & 63.56 \\
    TDA & \textit{CVPR2024} & 69.51 & 60.11 & 80.24 & 50.54 & 64.67 & 65.01 & 63.89 \\
    SCP & \textit{ACMMM2024} & 68.80 & 50.50 & 78.70 & 46.50 & 62.60 & 61.42 & 59.58 \\
    AdaPrompt & \textit{AAAI2024} & - & 47.71 & 73.98 & 47.72 & 59.32 & - & 57.18 \\
    C-TPT & \textit{ICLR 2024} & 69.30 & 52.90 & 78.00 & 48.50 & 63.40 & 62.42 & 60.70 \\
    O-TPT & \textit{CVPR2025} & 67.33 & 49.87 & 72.55 & 47.12 & 61.65 & 59.70 & 57.80 \\
    DynaPrompt & \textit{ICLR2025} & 69.61 & 56.17 & 78.17 & 48.22 & 64.67 & 63.37 & 61.81 \\
    TPS & \textit{WACV2025} & 71.45 & 60.61 & 80.20 & 50.88 & 64.91 & 65.61 & 64.15 \\
    \rowcolor{mygray}
    D\textsuperscript{2}TPT & \textit{This Paper} & \textbf{71.85} & \textbf{63.09} & \textbf{80.38} & \textbf{51.75} & \textbf{65.79} & \textbf{66.57} & \textbf{65.25} \\
    \bottomrule
    \end{tabular}
% }
\caption{Comparison of top-1 accuracy (\%) with baselines under natural distribution shifts. Best performances are in \textbf{bold}.}
\label{tab:d2tpt_dg}
\end{table*}

\subsection{Reliability-Aware Prompt Optimization}
Given \(\mathbf{L}_{RAM}^{r}\in\mathbb{R}^{M \times C}\), we obtain the corresponding learnable image feature vectors \(\mathbf{Z}_{img}^{r}\in \mathbb{R}^{M\times D}\). To further mitigate the impact of low-quality augmented views, we adopt a confidence-based weighted ensemble strategy to aggregate predictions from different augmented views. Specifically, we first compute the pairwise cosine similarity matrix \(\mathbf{S}=\hat{\mathbf{Z}}_{img}^{r}(\hat{\mathbf{Z}}_{img}^{r})^{T}\in \mathbb{R}^{M \times M}\), where \(\hat{\mathbf{Z}}_{img}^{r}\) denotes the normalized image features. The confidence score for each view is then obtained by summing its similarities with all other views, i.e., \(\textbf{s}=\left [\text{s}_{1};\cdots;\text{s}_{M}\right]\in\mathbb{R}^{M\times 1}\), where \(\text{s}_{i}=\sum_{j=1}^{M}\textbf{S}_{i,j}\). Therefore, we obtain the final ensemble weights \(\mathbf{w}=\text{softmax}(\textbf{s})\in \mathbb{R}^{M\times 1}\), which are subsequently used to compute the aggregated prediction \(\mathbf{L}_{EN}\in\mathbb{R}^{1\times C}\). Consequently, the corresponding self-entropy loss is computed as:
\begin{equation} \label{eq:d2tpt_15}
\mathcal{L}_{EN}=\mathcal{H}(\text{softmax}(\mathbf{L}_{EN})), 
\end{equation}
where \(\mathbf{L}_{EN}=\sum_{m=1}^{M}\mathbf{w}_{m}\cdot\mathbf{L}_{RAM}^{r,m}\), and \(\mathbf{w}_{m}\), \(\mathbf{L}_{RAM}^{r,m}\) represent the weight and prediction of the \(m\)-th augmented view, respectively.

To enhance the alignment between visual and textual modalities after incorporating learnable prompts, we propose a cross-modal consistency distillation strategy. Specifically, let \(\hat{\mathbf{Z}} _{\textit{text}}^{ori}\in\mathbb{R}^{C \times D}\) and \(\hat{\mathbf{Z}} _{\textit{text}}\in\mathbb{R}^{C \times D}\) denote the normalized text features before and after prompt adaptation, respectively. In addition, let \(\bar{\mathbf{Z}}_{img}^{r,ori}\in \mathbb{R}^{1\times D}\) and \(\bar{\mathbf{Z}}_{img}^{r}\in \mathbb{R}^{1\times D}\) represent the original and prompt-adapted image feature prototypes obtained after confidence selection. We define the cross-modal distillation logits as \(\mathbf{L}_{MD}\in \mathbb{R}^{1 \times C}\), and the corresponding self-entropy loss is defined as:
\begin{equation} \label{eq:d2tpt_17}
\mathcal{L}_{MD}= \mathcal{H}(\text{softmax}(\mathbf{L}_{MD})).
\end{equation}
where \(\mathbf{L}_{MD}=\mathbf{L}_{v\to t}+\mathbf{L}_{t\to v}+\mathbf{L}_{self}\), \(\mathbf{L}_{v\to t} = \exp{(s)} \cdot\left ( \bar{\mathbf{Z}}_{img}^{r,ori}(\hat{\mathbf{Z}} _{\textit{text}})^{T}  \right )\), \(\mathbf{L}_{t\to v} = \exp{(s)} \cdot\left ( \bar{\mathbf{Z}}_{img}^{r}(\hat{\mathbf{Z}} _{\textit{text}}^{ori})^{T}  \right )\), and \(\mathbf{L}_{self} = \exp{(s)} \cdot\left ( \bar{\mathbf{Z}}_{img}^{r}(\hat{\mathbf{Z}} _{\textit{text}})^{T}  \right )\).

By combining Equations (\ref{eq:d2tpt_13}), (\ref{eq:d2tpt_15}), and (\ref{eq:d2tpt_17}), we derive the final optimization objective:
\begin{equation} \label{eq:L_final}
    \boldsymbol{p_t}^{*}, \boldsymbol{p_v}^{*}=\arg\min_{\boldsymbol{p_t},\boldsymbol{p_v}}\mathcal{L},
\end{equation}
where \(\mathcal{L}=\mathcal{L}_{RAM}+\alpha \mathcal{L}_{EN}+\beta \mathcal{L}_{MD}\), and \( \alpha \), \( \beta \) are hyperparameters that balance the influence of \( \mathcal{L}_{EN} \) and \( \mathcal{L}_{MD} \), respectively. The procedural steps of D\textsuperscript{2}TPT are detailed in Algorithm~\ref{alg:d2tpt}.
% , respectively. 
% The detailed algorithmic description is provided in the Appendix.
% The training pipeline is detailed by Algorithm \ref{alg:ample_train}.

\section{Experiments}
In this section, we present experimental results to comprehensively evaluate the effectiveness of D\textsuperscript{2}TPT.
% In this section, we present experimental results to comprehensively evaluate the effectiveness and robustness of D\textsuperscript{2}TPT in comparison to existing baselines.

\subsection{Experimental Setup}

\textbf{Datasets.} Following prior work~\citep{DBLP:conf/nips/ShuNHYGAX22}, we evaluate our method under two key scenarios: (1) Natural distribution shifts. We use ImageNet~\citep{DBLP:conf/cvpr/DengDSLL009} and four of its out-of-distribution variants, i.e., ImageNet-A~\citep{DBLP:conf/cvpr/HendrycksZBSS21}, ImageNet-R~\citep{DBLP:conf/iccv/HendrycksBMKWDD21}, ImageNet-Sketch~\citep{DBLP:conf/nips/WangGLX19}, and ImageNet-V2~\citep{DBLP:conf/icml/RechtRSS19}. (2) Cross-datasets generalization. We conduct evaluations across ten diverse classification datasets, including Caltech101~\citep{DBLP:conf/cvpr/LiFP04}, OxfordPets~\citep{DBLP:conf/cvpr/ParkhiVZJ12}, StanfordCars~\citep{DBLP:conf/iccvw/Krause0DF13}, Flowers102~\citep{DBLP:conf/icvgip/NilsbackZ08}, Food101~\citep{DBLP:conf/eccv/BossardGG14}, Aircraft~\citep{DBLP:journals/corr/MajiRKBV13}, SUN397~\citep{DBLP:conf/cvpr/XiaoHEOT10}, DTD~\citep{DBLP:conf/cvpr/CimpoiMKMV14}, EuroSAT~\citep{DBLP:journals/staeors/HelberBDB19}, and UCF101~\citep{DBLP:journals/corr/abs-1212-0402}.
% These datasets serve as a comprehensive benchmark for assessing model robustness under varied distributional conditions. 
Table~\ref{tab:dataset-overview} presents the detailed statistics of these datasets.

\textbf{Baselines.} We compare our D\textsuperscript{2}TPT against a comprehensive set of baselines, including both zero-shot and test-time adaptation approaches built upon the CLIP ViT-B/16~\citep{DBLP:conf/icml/RadfordKHRGASAM21} backbone. Specifically, we consider the following methods: TPT~\citep{DBLP:conf/nips/ShuNHYGAX22}, DiffTPT~\citep{DBLP:conf/iccv/Feng0LKZ23}, PromptAlign~\citep{DBLP:conf/nips/SamadhGHKNK023}, TDA~\citep{DBLP:conf/cvpr/KarmanovGLEX24}, SCP~\citep{DBLP:conf/mm/WangZ0024}, AdaPrompt~\citep{DBLP:conf/aaai/Zhang0L24}, C-TPT~\citep{DBLP:conf/iclr/YoonYTHLY24}, O-TPT~\citep{DBLP:conf/cvpr/SharifdeenMBKK25}, DynaPrompt~\citep{DBLP:conf/iclr/XiaoYHCJHSWS25}, and TPS~\citep{DBLP:conf/wacv/SuiWY25}. These baselines encompass a diverse range of strategies for adapting VLMs at test time, including prompt tuning, distribution alignment, and other robust adaptation techniques. 
% Our evaluation is conducted with a consistent backbone and evaluation benchmark, highlighting the effectiveness and generalization capability of D\textsuperscript{2}TPT.

% BCA~\citep{DBLP:conf/cvpr/Zhou0LLZDLL25}

\begin{table*}[t]
    \centering
    % \vspace{-3mm}
    \setlength{\tabcolsep}{1.8mm}
    % \adjustbox{max width=\textwidth}{
    \begin{tabular}{l l lllllllllll}
    \toprule
    \rotatebox{90}{Method} & \rotatebox{90}{Publication} & \rotatebox{90}{Caltech101} & \rotatebox{90}{OxfordPets} & \rotatebox{90}{StanfordCars} & \rotatebox{90}{Flowers102} & \rotatebox{90}{Food101} & \rotatebox{90}{Aircraft} & \rotatebox{90}{SUN397} & \rotatebox{90}{DTD} & \rotatebox{90}{EuroSAT} & \rotatebox{90}{UCF101} & \rotatebox{90}{\emph{Average}} \\
    \midrule
    CLIP-ViT-B/16 & \textit{ICML2021} & 93.35 & 88.25 & 65.48 & 67.44 & 83.65 & 23.67 & 62.59 & 44.27 & 42.01 & 65.13 & 63.58 \\
    \midrule
    TPT & \textit{NeurIPS2022} & 94.16 & 87.79 & 66.87 & 68.98 & 84.67 & 24.78 & 65.50 & 47.75 & 42.44 & 68.04 & 65.10 \\
    DiffTPT & \textit{ICCV2023} & 92.49 & 88.22 & 67.01 & 70.10 & 87.23 & 25.60 & 65.74 & 47.00 & 43.13 & 68.22 & 65.47 \\
    PromptAlign & \textit{NeurIPS2023} & 94.01 & \textbf{90.76} & 68.50 & 72.39 & 86.65 & 24.80 & 67.54 & 47.24 & 47.86 & 69.47 & 66.92  \\
    TDA & \textit{CVPR2024} & 94.24 & 88.63 & 67.28 & 71.42 & 86.14 & 23.91 & 67.62 & 47.40 & \textbf{58.00} & 70.66 & 67.53  \\
    SCP & \textit{ACMMM2024} & 93.90 & 88.60 & 65.90 & 70.00 & \textbf{87.40} & 24.80 & 69.10 & 43.90 & 47.30 & 67.80 & 65.87 \\
    AdaPrompt & \textit{AAAI2024} & 94.07 & 89.64 & 63.29 & 72.97 & 84.72 & 21.21 & 65.37 & 44.75 & 47.20 & 67.22 & 65.04 \\
    C-TPT & \textit{ICLR 2024} & 94.10 & 87.40 & 66.70 & 69.90 & 84.50 & 23.90 & 66.00 & 46.80 & 48.70 & 66.70 & 65.47 \\
    O-TPT & \textit{CVPR2025} & 93.95 & 87.95 & 64.53 & 70.07 & 84.13 & 23.64 & 64.23 & 45.68 & 42.84 & 64.16 & 64.12 \\
    DynaPrompt & \textit{ICLR2025} & 94.32 & 88.28 & 67.65 & 69.95 & 85.42 & 24.33 & 66.32 & 47.96 & 42.28 & 68.72 & 65.52 \\
    TPS & \textit{WACV2025} & 95.09 & 87.35 & 69.06 & 71.54 & 85.23 & 26.34 & 68.98 & 50.47 & 44.48 & 71.00 & 66.95 \\
    \midrule
    \rowcolor{mygray}
    D\textsuperscript{2}TPT & \textit{This Paper} & \textbf{95.10} & 87.55 & \textbf{69.50} & \textbf{75.21} & 84.95 & \textbf{27.00} & \textbf{69.80} & \textbf{52.88} & 54.92 & \textbf{72.40} & \textbf{68.93} \\
    \bottomrule
    \end{tabular}
    % }
    % \vspace{-2mm}
    \caption{Comparison of top-1 accuracy (\%) with baselines under cross-datasets generalization. Best performances are in \textbf{bold}.}
    \label{tab:d2tpt_xd} %\vspace{-1em}
\end{table*}

\textbf{Implementation Details.} We set the register capacity \(K\) for each class in the knowledge base to 3. Following TPT~\citep{DBLP:conf/nips/ShuNHYGAX22}, each test image is augmented 63 times via random resized cropping and grouped with the original into a batch of 64. Among the predictions, we select the top 10\% (\(\tau = 0.1\)) most confident samples and compute the entropy of their averaged probability. The learnable prompts are then updated for one step by minimizing this entropy using the AdamW optimizer. All experiments are conducted on a single NVIDIA GeForce RTX 4090, and results are averaged over three random seeds.

\begin{table}[t]
    \centering
    % \vspace{-3mm}
    % \resizebox{0.9\columnwidth}{!}{
    \setlength{\tabcolsep}{4.2mm}
    \begin{tabular}{ccccc}
    \toprule
     RAM & \multicolumn{2}{c}{RPO} & NDS & CDG \\
      & CWE & CMD & & \\
    \midrule
    $\times$ & $\times$ & $\times$ & 64.57 & 67.08 \\
    $\checkmark$ & $\times$ & $\times$ & 64.82 & 68.27 \\
    $\checkmark$ & $\checkmark$ & $\times$ & 65.02 & 68.43 \\
    $\checkmark$ & $\checkmark$ & $\checkmark$ & 65.25 & 68.93 \\
    
    \bottomrule
    \end{tabular}
    % }
    % \vspace{-2mm}
    \caption{Ablation study on the effectiveness of each module. NDS and CDG are abbreviations for natural distribution shifts and cross-datasets generalization, respectively.}
    \label{tab:d2tpt_ablation}
\end{table}

\begin{figure}[t]
\centering
\includegraphics[width=0.46\textwidth]{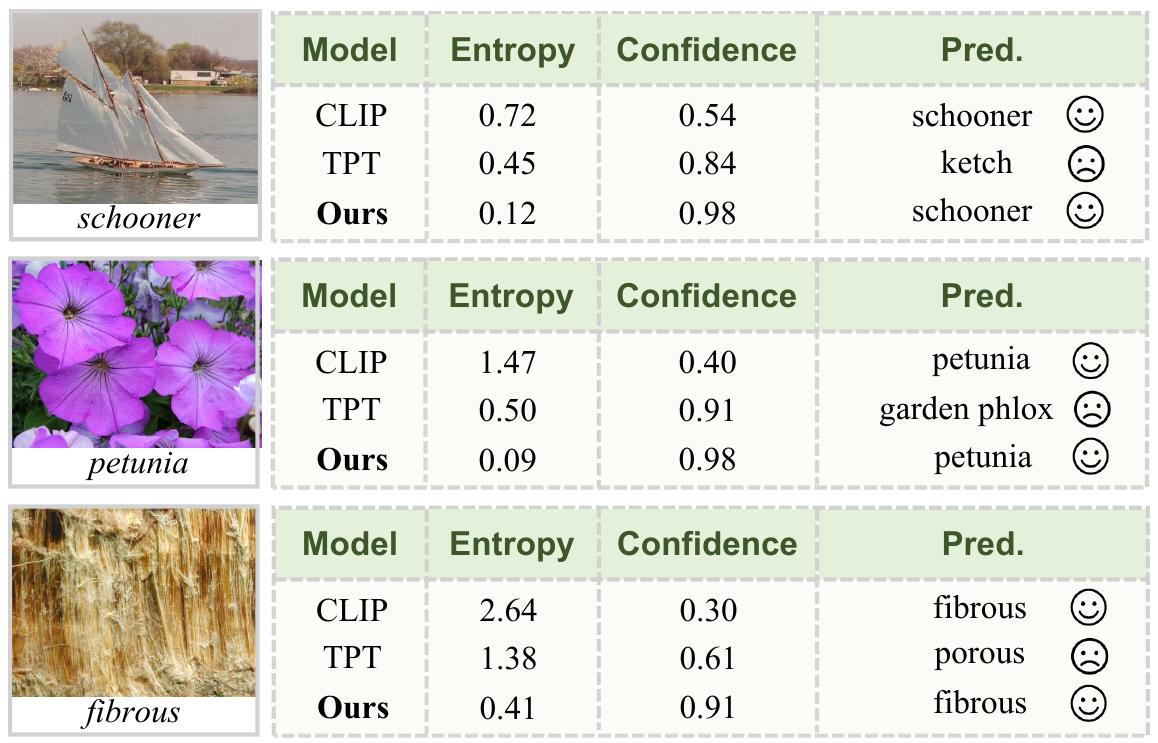} % Reduce the figure size so that it is slightly narrower than the column.
\caption{Case study on prediction confidence and correctness. We use pred. to denote the model’s predicted category.}
\label{fig:case_study}
\end{figure}

\subsection{Main Results}
\textbf{Natural Distribution Shifts.} Table~\ref{tab:d2tpt_dg} presents a comparison of top-1 accuracy across various test datasets under natural distribution shifts. Our proposed method, D\textsuperscript{2}TPT, consistently outperforms all baselines. Specifically, it achieves the highest average accuracy of 66.57\%, outperforming all competitors and demonstrating superior generalization capability. In terms of OOD average accuracy, D\textsuperscript{2}TPT reaches 65.25\%, surpassing the primary baseline TPS by 1.1\%. These results validate the effectiveness of our D\textsuperscript{2}TPT in handling distribution shifts across diverse visual domains.

\textbf{Cross-Datasets Generalization.} Table~\ref{tab:d2tpt_xd} reports the top-1 accuracy on 10 diverse datasets for evaluating cross-datasets generalization. Our method, D\textsuperscript{2}TPT, achieves the highest average accuracy of 68.93\%, outperforming all previous baselines. In particular, D\textsuperscript{2}TPT obtains the best results on 7 out of 10 datasets. Compared to the primary baseline TPS, D\textsuperscript{2}TPT brings a performance gain of 1.98\% on average. These results clearly demonstrate the generalization capability of D\textsuperscript{2}TPT across diverse visual domains with varying distributions and label semantics.

\textbf{Ablation Study.} We assess the impact of removing two modules, i.e., dynamic retrieval-augmented modulation (RAM) and reliability-aware prompt optimization (RPO), on model performance. CWE (confidence-based weighted ensemble) and CMD (cross-modal consistency distillation) are subcomponents of the RPO module. As shown in Table~\ref{tab:d2tpt_ablation}, the experimental results highlight the individual contributions of each module. Specifically, removing either the RAM or RPO module leads to a performance drop, while the model achieves the best performance when both modules are used jointly. This confirms the effectiveness of each component in our proposed approach.

\begin{table}[t]
    \centering
    % \vspace{-3mm}
    % \resizebox{0.9\columnwidth}{!}{
    \setlength{\tabcolsep}{5.0mm}
    \begin{tabular}{l cc}
    \toprule
    & NDS & CDG \\
    \midrule
    D\textsuperscript{2}TPT vs. TPS & 0.046 & 0.032 \\
    D\textsuperscript{2}TPT vs. DynaPrompt & 0.033 & 0.021 \\
    D\textsuperscript{2}TPT vs. O-TPT & 0.016 & 0.003 \\
    \bottomrule
    \end{tabular}
    % }
    % \vspace{-2mm}
    \caption{The \textit{p}-value for Student's \textit{t}-test across two settings.}
    \label{tab:t_test}
\end{table}

\begin{figure}[t]
\centering
\includegraphics[width=0.46\textwidth]{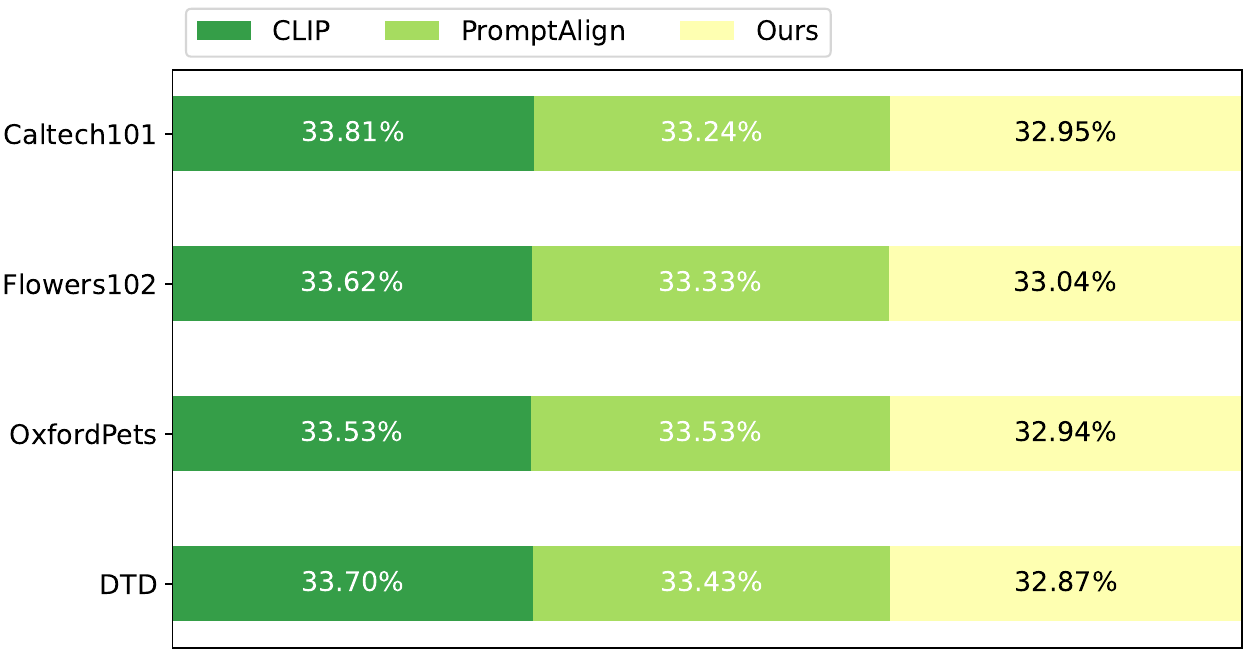} % Reduce the figure size so that it is slightly narrower than the column.
\caption{Comparison of normalized cross-modal feature distances across different models.}
\label{fig:distance_experiments}
\end{figure}

\begin{table}[t]
    \centering
    % \vspace{-3mm}
    % \resizebox{0.9\columnwidth}{!}{
    \setlength{\tabcolsep}{2.2mm}
    \begin{tabular}{cccc}
    \toprule
    Method & Top-1 Accuracy & Params & FPS  \\
    \midrule
    {TPT} & 68.98 & 2048 & 6.37 \\
    {PromptAlign} & 72.39 & 1185024 & 7.41 \\
    {TPS} & 71.54 & 52224 & 6.02   \\
    {D\textsuperscript{2}TPT} & 75.21 & 52736 & 5.71 \\
    \bottomrule
    \end{tabular}
    % }
    % \vspace{-2mm}
    \caption{Complexity comparison of D\textsuperscript{2}TPT and baselines on the Flowers102 dataset.}
    \label{tab:d2tpt_complexity}
\end{table}

\subsection{Further Analysis}
\textbf{Case Study.} Figure~\ref{fig:case_study} presents representative examples where TPT tends to produce low-entropy yet incorrect predictions, reflecting overconfident misclassification. In contrast, our method effectively suppresses such overconfident errors, which can be attributed to the design strategy that modulates the original model predictions during prompt optimization. These examples suggest that our approach mitigates the prompt optimization bias, enabling more confident and accurate predictions.

\textbf{Significance Test.} To verify that the observed performance improvements are not attributable to random chance, a Student’s t-test~\citep{mendenhall2020introduction} is conducted between the baseline models and the proposed D\textsuperscript{2}TPT. A p-value below 0.05 is considered statistically significant. As reported in Table~\ref{tab:t_test}, all p-values across the two task settings, i.e., natural distribution shifts and cross-datasets generalization, fall below this threshold. Specifically, the comparisons between TPS and D\textsuperscript{2}TPT yield p-values of 0.046 and 0.032, respectively. These results indicate that D\textsuperscript{2}TPT achieves statistically significant improvements over the baselines, further validating the effectiveness of the proposed method.

\textbf{Computational Complexity.} To further assess the trade-off between accuracy and computational efficiency, we conduct a complexity analysis under the cross-datasets generalization setting, with results summarized in Table~\ref{tab:d2tpt_complexity}. On the Flowers102 dataset, D\textsuperscript{2}TPT achieves a Top-1 accuracy of 75.21\% with 52736 trainable parameters and an inference throughput of 5.71 FPS. These results are comparable to those of the baseline models. Although D\textsuperscript{2}TPT exhibits a slightly lower inference throughput, its relatively low parameter count results in only a minimal increase in computational complexity, demonstrating that the improvements in accuracy come with negligible computational overhead.

\textbf{Cross-Modal Alignment Evaluation.} 
% To further assess the effectiveness of our method in aligning visual and textual modalities, we compute the L2 distances between image and text features across four benchmark datasets: Caltech101, Flowers102, OxfordPets, and DTD. 
We compute the L2 distances between image and text features to further assess the effectiveness of our method in aligning visual and textual modalities. To facilitate clearer comparisons among models within each dataset, the distances are normalized as proportions. The normalized results are shown in Figure~\ref{fig:distance_experiments}, where smaller proportions indicate shorter distances and thus better cross-modal alignment. We observe that our method consistently achieves the lowest proportion of feature distance, outperforming both CLIP and PromptAlign, which indicates a stronger alignment between modalities. 
% These results suggest that our method enhances the semantic consistency between visual and textual representations, thereby facilitating more reliable cross-modal understanding.

\begin{figure}[t]
\centering
\includegraphics[width=0.45\textwidth]{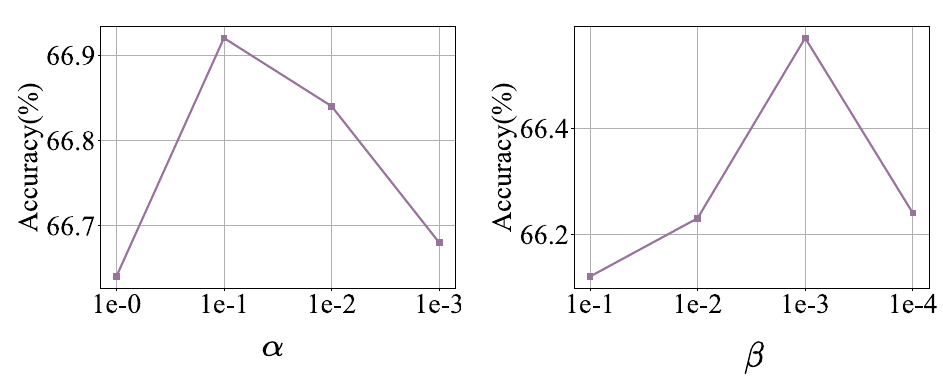} % Reduce the figure size so that it is slightly narrower than the column.
\caption{Influence of \(\alpha\) and \(\beta\) on model accuracy.}
\label{fig:hyper-param}
\end{figure}

\textbf{Hyperparameter Analysis.} To systematically investigate the impact of the hyperparameters \(\alpha\) and \(\beta\) on model performance, we conduct an empirical study under the cross-datasets generalization setting. Specifically, we search for the optimal value of \(\alpha\) from the set \(\{1e^{0}, 1e^{-1}, 1e^{-2}, 1e^{-3}\}\) and \(\beta\) from \(\{1e^{-1}, 1e^{-2}, 1e^{-3}, 1e^{-4}\}\), using four representative datasets: Caltech101, Flowers102, UCF101, and Aircraft. Figure~\ref{fig:hyper-param} presents the average performance across the four datasets under different settings of the two hyperparameters. Based on the results, we select \(\alpha = 1e^{-1}\) and \(\beta = 1e^{-3}\) as the final configuration for our model.

\section{Conclusion}
In this work, we propose Doubly Debiased Test-Time Prompt Tuning (D\textsuperscript{2}TPT) to mitigate prompt optimization bias from both model and data perspectives. By incorporating a dynamic retrieval-augmented modulation module and a reliability-aware prompt optimization module, D\textsuperscript{2}TPT enables more confident and accurate predictions, along with stronger alignment between visual and textual modalities. These improvements jointly alleviate prompt optimization bias and enhance the performance of test-time prompt tuning. Extensive experiments demonstrate that D\textsuperscript{2}TPT consistently outperforms existing baselines on benchmarks involving natural distribution shifts and cross-datasets generalization, validating its effectiveness and robustness.

\section{Acknowledgments}
This work is supported by the National Natural Science Foundation of China No. 62406313, 2023 Special Research Assistant Grant Project of the Chinese Academy of Sciences.

\bibliography{aaai2026}

\end{document}